\documentclass[twoside,twocolumn,10pt]{article}

\usepackage{wscg}           
\RequirePackage{ifpdf}
\ifpdf
 \RequirePackage[pdftex]{graphicx}
 \RequirePackage[pdftex]{color}
\else
 \RequirePackage[dvips,draft]{graphicx}
 \RequirePackage[dvips]{color}
\fi
\usepackage[german,english]{babel}     
\usepackage{array}          
\usepackage{tabularx}      
\usepackage{nopageno}       
\usepackage{amssymb}
\usepackage{amsmath}
\usepackage{multicol}
\usepackage{multirow}
\usepackage{xspace}
\title{Dynamic Sensor Matching based on\\ Geomagnetic Inertial Navigation}
\author{
\hspace{-1.3cm}
\parbox{0.45\textwidth}{\centering
Simone M{\"u}ller\\[1mm]
Leibniz Supercomputing Centre (LRZ)\\[1mm]
Boltzmannstrasse 1\\
85748 Garching bei M{\"u}nchen\\[1mm]
simone.mueller@lrz.de}
\hspace{0.5cm}
\parbox{0.45\textwidth}{\centering
Dieter Kranzlm{\"u}ller\\[1mm]
Ludwig-Maximilians-Universit{\"a}t (LMU)\\[1mm]
MNM-Team \\
Oettingenstr. 67 \\
80538 M{\"u}nchen \\ [1mm]
kranzlmueller@ifi.lmu.de}
}
\usepackage{url}
\makeatletter
\def\Uslash{\mathbin{\mathchar`\/}\@ifnextchar{/}{\kern-.15em}{}}
\g@addto@macro\UrlSpecials{\do \/ {\Uslash}}
\def\Ucolon{\mathbin{\mathchar`:}\@ifnextchar{/}{\kern-.1em}{}}
\g@addto@macro\UrlSpecials{\do : {\Ucolon}}
\makeatother
\begin{document}
\twocolumn[{
\csname @twocolumnfalse\endcsname
\maketitle  
\begin{abstract}
\noindent

Optical sensors can capture dynamic environments and derive depth information in near real-time. The quality of these digital reconstructions is determined by factors like illumination, surface and texture conditions, sensing speed and other sensor characteristics as well as the sensor-object relations. Improvements can be obtained by using dynamically collected data from multiple sensors. However, matching the data from multiple sensors requires a shared world coordinate system. We present a concept for transferring multi-sensor data into a commonly referenced world coordinate system: the earth's magnetic field. The steady presence of our planetary magnetic field provides a reliable world coordinate system, which can serve as a reference for a position-defined reconstruction of dynamic environments. Our approach is evaluated using magnetic field sensors of the ZED 2 stereo camera from Stereolabs, which provides orientation relative to the North Pole similar to a compass. With the help of inertial measurement unit informations, each camera's position data can be transferred into the unified world coordinate system. Our evaluation reveals the level of quality possible using the earth magnetic field and allows a basis for dynamic and real-time-based applications of optical multi-sensors for environment detection.

\end{abstract}
\subsection*{Keywords}
Dynamic Matching, Multi-Sensor, Magnetic Inertial Navigation, Real-Time, Computer Vision
\vspace*{1.0\baselineskip}
}]
\section{Introduction}
\copyrightspace
 
Many devices use 3D reconstructions of their surroundings for locomotion and interaction in complex visual environments \cite{kok18}. Epipolar geometry based distance information of depth sensors allows us to compute 3D points as point clouds. Multiple depth sensors can be used efficiently for real-time point cloud expanding and optimising \cite{pia13,mue21}. The positional overlay of received depth information reduces sensor and image errors. This involves the positional accuracy and stability of depth sensors. Using global navigation of inertial navigation systems (GNSS/INS) enables a temporal position adjustment of these sensors \cite{hua19,vu12}. However, tracking systems like Garmin Oregon 700 are insufficient for the matching of multiple sensors due to positional deviations of 3~m to 25~m \cite{gar22}. These position deviations affect the quality of rendered point clouds due to incorrect coordinate alignments, scattering, outliers or offsets of neighbouring depth points \cite{car16,kad14}. The required visual contact between satellite and receiver additionally limits the local GPS signal \cite{chu11}. Alternative Visual Inertial Navigation Systems (VINS) offer limited benefits for multiple sensor matching in terms of sensor drifts, measurement errors and range \cite{car16,hua19,kad14}.

Our motivation is based on the challenges of using stable inertial navigation systems for multiple sensor matching. The persistence of the geomagnetic field lends itself to our concept. We apply the combination of 3D depth technologies and smart sensor architectures for the use of magnetic fields in an inertial system. Smart sensor architectures offer a gradient transformation between acceleration, angular velocity and magnetic field in the meter defined world coordinate system $W_\mathrm{(x,y,z)}$ \cite{han07,car16}. Our contribution comprises the following aspects:
\vspace*{-0.1cm}
\begin{itemize}
    \item Location and time independent matching of multiple sensors based on geomagnetic inertial navigation \vspace*{-0.2cm}
    \item Sensory setup for validation of our approach \vspace*{-0.2cm}
    \item Analysis of positioning-based accuracy, reliability and stability in a transformed world coordinate system 
\end{itemize}

\begin{center}
\begin{table}
\begin{tabular}{r l }
\hline
\textbf{Notation}	& \textbf{Definition}	\hspace*{2cm}    \\
\hline
& \vspace*{-0.36cm}\\
$a$ & Acceleration [$m/s^{2}$] \\
$g$ & Gravity 9.81 [$m/s^{2}$] \\
$B$ & Baseline between $C_\mathrm{L}$ and $C_\mathrm{R}$ [cm] \\
$B_\mathrm{1}$, $B_\mathrm{n}$ & Magnetic field strength [$\mu T$]\\
$b_\mathrm{H}$   & Temperature Dependent Bias [$\mu T$]	\vspace*{0.01cm} \\
$C$   & Transformation Matrix [$\mu T$]	\vspace*{0.01cm} \\
$m$ & Seismic mass of accelerometer [$mg$] \\
$R_\mathrm{(\psi,\phi,\theta)}$    &  Rotation (Pitch$\psi$, Roll$\phi$, Yaw$\theta$) \vspace*{0.01cm} \\
$s$   & Distance [$m$]	\vspace*{0.01cm} \\
$t$   & Time [$s$]	\vspace*{0.01cm} \\
$T$   & Translation [$m$]	\vspace*{0.01cm} \\
$v$   & Velocity [$m/s²$]	\vspace*{0.01cm} \\
$\lambda$  & Depth [$m$] \vspace*{0.00cm} \\
$\varphi$ & Rotation angle [$rad$] \\
$\theta$ & Orientation [$^{\circ}$]\\
$\omega$ & Angular velocity [$rad/s$] \\
& \vspace*{-0.36cm}\\
\hline
\textbf{Abbr.} & \textbf{Definition} \\
\hline
& \vspace*{-0.36cm}\\
CMOS & Complementary Metal \\
&  Oxide Semiconductor \\
GNSS & Global Navigation Satellite Systems \\
GPS & Global Positioning System \\
ICP & Iterative Closest Point \\
IMU & Inertial Measurement Unit \\
INS & Inertial Navigation Systems \\
LiDAR &  Light Detection And Ranging \\
SLAM & Simultaneous Localisation and Mapping \\
ToF & Time of Flight \\
VO & Visual Odometry \\
VINS & Visual Inertial Navigation Systems \\
VIO & Visual Inertial Odometry \\
WCS & World Coordinate System \\
\hline
\end{tabular}
\caption{\textbf{Notations and Abbreviations}}
\label{fig:Notations}
\end{table}
\end{center}

\vspace*{-0.5cm}


Table~\ref{fig:Notations} lists the used definitions, notations and abbreviations. This Paper is organised as follows:
\begin{itemize}
    \item \textbf{Section 2} overviews related works of sensor matching and valid transformations in inertial systems. \vspace*{-0.25cm}
    \item \textbf{Section 3} discusses our concept of geomagnetic inertial navigation and coordinate transformation of the magnetic field sensor. \vspace*{-0.25cm}
    \item \textbf{Section 4} lists the steps of our methodology for the evaluation. \vspace*{-0.25cm}
    \item \textbf{Section 5} overviews setup information of our measurement implementation. \vspace*{-0.25cm}
    \item \textbf{Section 6} discusses the static and dynamic results of our geomagnetic inertial navigation.\vspace*{-0.25cm}
    \item \textbf{Section 7} describes our conclusion and resulting future work.
\end{itemize}

\section{Related Work}
The challenges of dynamic matching demands a referencable and accurate world coordinate system (WCS) in which several depth sensors move in defined positions. In this section, we describe the matching of multiple sensors and transferability to this WCS. Based on related work, we provide a foundation for our hypothesis and methodological approach.

\textbf{Multiple Depth Sensor Systems} Locomotive sensors require information about their translation $T_\mathrm{x,y,z}$, rotation $R_\mathrm{\psi,\phi,\theta}$ and local depth $\lambda$ \cite{mue21}.
\begin{figure}[ht!]
    \centering
    \includegraphics[scale=0.25]{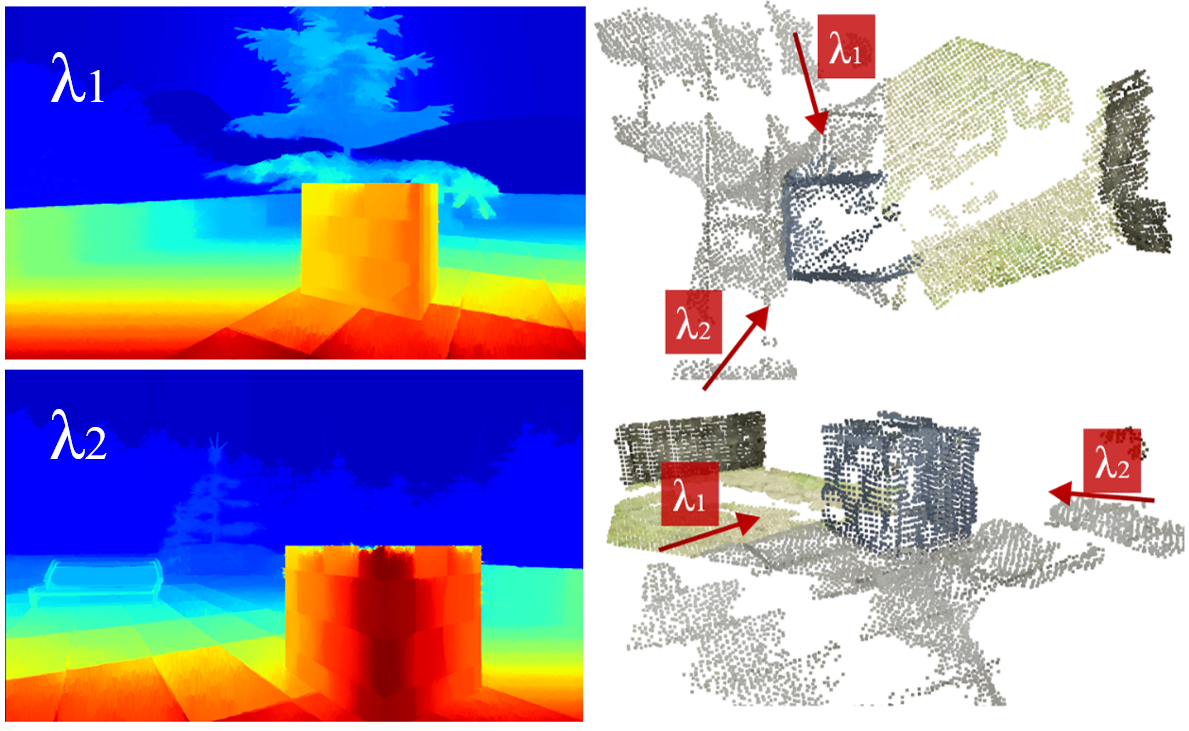}
    \caption{\textbf{Point Cloud Matching}: The resulting point cloud (right) is matched from several depth images (top and bottom left). The arrows of $\lambda_\mathrm{1}$ and $\lambda_\mathrm{2}$ represent the camera orientation. }
    \label{fig:SensorMatching}
\end{figure}

3D depth technologies like Light Detection And Ranging (LiDAR), Time of Flight (ToF) or stereoscopic systems receive the emitted light of their surroundings and convert it to electrical signals \cite{yoe21}. The depth information ($\lambda_\mathrm{1}$ and $\lambda_\mathrm{2}$ on the left side of Fig.~\ref{fig:SensorMatching}) can be calculated algorithmically from the converted digital sensor signals. One commonly used algorithm for point cloud registration is Iterative Closest Point (ICP), which involves the determination of point cloud specific rotation and translation \cite{tar10, mue21}. Approaches like Simultaneous Localisation and Mapping (SLAM), Visual Odometry (VO), visual detection and tracking, as well as visual classification and recognition enable 3D evaluations by means of volumetric rendered point clouds (right side of Fig.~\ref{fig:SensorMatching}) \cite{mue21}. Volumetric data can be converted into polygonal meshes in order to manipulate objects volumetrically on the fly \cite{tak15}.

Quality characteristics of these point clouds can be increased by matching multiple depth images as shown in Fig.~\ref{fig:SensorMatching}. Error corrections of distortions, scatter, noise, sensor defects or latencies as well as point cloud deviations characterise such quality features \cite{kad14,mue21,tak15}.

Takimotoa et al. \cite{tak15} examined the matching of multiple point clouds and described the finding of ICP correspondences as challenges when texture-free surfaces of point clouds are to be reconstructed densely and accurately. Their measurements indicate that low-precision sensors are capable of reasonably good reconstructed objects. They conclude that increased number of acquisitions and SIFT methodology do not improve the final point cloud quality. Instead, they suggest that the adjustment of ICP and reconstruction parameters leads to improvements due to limited sensor accuracy and stability.

Piatkowska et al. \cite{pia13} optimised spatio-temporal and three-dimensional reconstructions with asynchronous time-based image sensors and extended existing methods for event-based processing. They conclude that dynamic, asynchronous and cooperative implementations are possible using a specialised algorithm.

In \cite{mue21}, we describe the possibility of synchronous and dynamic sensor matching for limited sensor and object ranges. We name the alignment and transferability relevance of precise sensor positions for successful depth matching. 

\vspace*{-0.3cm}
\begin{figure}[ht!]
    \centering
    \hspace*{-0.2cm} \includegraphics[scale=0.4]{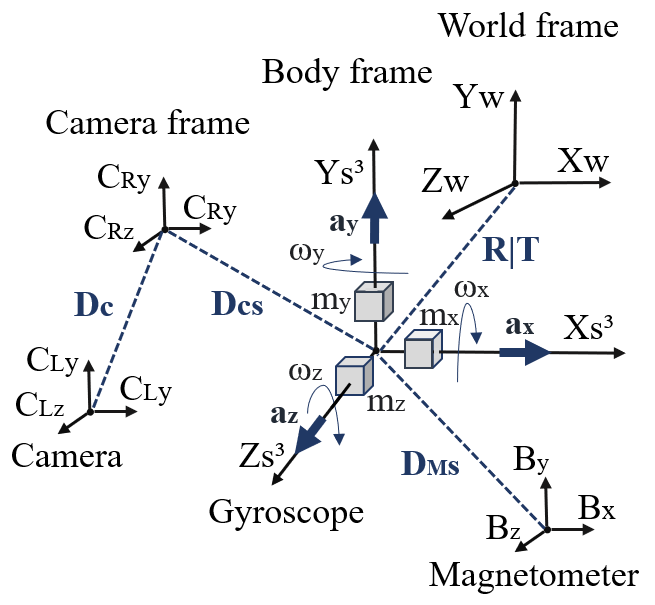}
    \caption{\textbf{Combination of 3D Depth Sensing and Smart Sensor Architecture}: Relationships between the stereo camera and IMU integrated gyroscope and magnetometer in a defined world frame coordinate system. The received images of $C_\mathrm{R}$ and $C_\mathrm{L}$ are located in the extrinsic camera frame \cite{mue21}. In contrast to the cameras, the IMU is originally located in body frame.}
    \label{fig:SensorCombination}
\end{figure}
\textbf{Inertial Navigation} Prior research indicates that minor deviations of aligned point clouds can be minimised by accurate positional information. Data fusion of inertial sensors in form of smart sensor architectures (see Fig.~\ref{fig:SensorCombination}) reduce the influence of a failure prone single sensor \cite{car16}. This allows the necessary optimisation of stability and precise movement detection.

 Inertial Measurement Unit (IMU) and depth integrated sensors have statically defined distances $D_\mathrm{CS}$, $D_\mathrm{MS}$ and $D_\mathrm{c}$ to each other. Estimated sensor positions and orientations are coordinated in camera extrinsic depth sensors or body frame IMUs \cite{car16,mue21}. Positional relations of multiple used sensors are not directly transferable into the world frame since each sensor uses its own coordinate definitions. Therefore, rotations and translations have to be transformed from camera and body into a world frame defined coordinate system \cite{car16}. The attitude representation of quaternions $q$ as shown in Eq.~\ref{eq:Quatrion} describes Euler's principle rotations $\varphi$ from inertial to body frame \cite{vec22,car16}. We obtain the current position by measuring actual speed $v$, angular velocity $\omega$, acceleration $a$, gravity $g$, and local magnetic field $B$ (Eq.~\ref{eq:QuatrionAngularVelocityV1},~\ref{eq:QuatrionAngularVelocityV2} and ~\ref{eq:QuatrionMagneticFieldV1}) \cite{car16}.
\vspace*{-0.0cm}
\begin{equation}\label{eq:Quatrion}
     q = [cos \dfrac{\varphi}{2}, \varphi \cdot sin \dfrac{\varphi}{2}]^{T} \hspace*{0.0cm}
\end{equation}
\begin{equation}\label{eq:QuatrionAngularVelocityV1}
    \dfrac{\partial q}{\partial t} = \dfrac{1}{2} q \cdot \omega \hspace*{1.9cm}
\end{equation}
\vspace*{-0.3cm}
\begin{equation}\label{eq:QuatrionAngularVelocityV2}
    \dfrac{\partial v}{\partial t} = \omega \times v + a + q \cdot g \cdot q^{1}
\end{equation}
\begin{equation}\label{eq:QuatrionMagneticFieldV1}
    \dfrac{\partial B}{\partial t} = \omega \times B + \nabla B \cdot v  \hspace*{0.7cm}
\end{equation}

We use $\nabla$ as $3\times3$ gradient matrix. A continuous and time-dependent initialisation of body oriented dynamic sensors in the transformed world frame is necessary to ensure positional accuracy.

Smart sensor architectures include IMUs consisting of three accelerometers, gyroscopes and magnetometer \cite{auf11}. The body frame denotated angular velocity $\omega$ of the gyroscope (Eq.~\ref{eq:GyroscopeVelocity}) can be determined from the measured angular velocity $\omega_\mathrm{m}$, temperature dependent bias $b_\mathrm{t}$ and additive $\eta$ of zero-mean Gaussian noise \cite{vec22,kok18}.
\begin{equation}\label{eq:GyroscopeVelocity}
    Gyroscope: \omega = \omega_\mathrm{m} + b_\mathrm{t} + \eta \hspace*{0.1cm}\{\hspace*{0.1cm} \eta \sim N (0,\sigma_\mathrm{gyro}^{2})
\end{equation}
$\omega_\mathrm{m}$ defines the sensor's angular velocity of body frame with respect to earth inertial frame \cite{kok18}.
\begin{equation}\label{eq:GyroscopeOrientation}
    Orientation: \theta_\mathrm{(t + \Delta t)} \approx \theta_\mathrm{(t)} + \dfrac{\partial}{\partial t} \theta_\mathrm{(t)} \Delta t + \epsilon
\end{equation}
The orientation $\theta$ (Eq.~\ref{eq:GyroscopeOrientation}) of gyroscope measurements can be approximated by Taylor expansion \cite{vec22}. $\theta_\mathrm{(t + \Delta t)}$ describes the angle at current step and $\theta_\mathrm{(t)}$ defines last changed time step $\Delta t$. $\epsilon$ is the approximation error. $R$ is presented by Euler angles ($\phi,\theta,\psi$) \cite{kok18,mue21}. 
 \begin{equation}\label{eq:Roll}
    R_\mathrm{x(\phi)} = 
    \left(
\begin{array}{c c c}
    1 & 0 & 0\\
    0 & cos\phi & -sin\phi\\ 
    0 & sin\phi & cos\phi\\  
\end{array}
\right)
\end{equation}
\begin{equation}\label{eq:Yaw}
    R_\mathrm{y(\theta)} = 
    \left(
\begin{array}{c c c}
    cos\theta & 0 & sin\theta\\
    0 & 1 & 0\\ 
    -sin\theta & 0 & cos\theta\\  
\end{array}
\right)
\end{equation}
\begin{equation}\label{eq:Pitch}
    R_\mathrm{z(\psi)} = 
  \left(
\begin{array}{c c c}
    cos\psi & -sin\psi & 0\\
    sin\psi & cos\psi & 0\\ 
    0 & 0 & 1\\  
\end{array}
\right)
\end{equation}
We can calculate the translation by measuring the linear acceleration $a_\mathrm{lin}$ (Eq.~\ref{eq:Accelerometer}) of accelerometer (Acc) \cite{kok18}. 
\begin{equation}\label{eq:Accelerometer}
  Acc:  a_\mathrm{lin} = a^{(g)} + a^{(l)} + \eta \hspace*{0.1cm}\{\hspace*{0.1cm} \eta \sim N (0,\sigma_\mathrm{acc}^{2})
\end{equation}
In motionless state, we measure the noisy gravity vector $a^{(g)}$ and zero-mean Gaussian noise $\eta$ with a magnitude of $9.81 m/s^{2} = 1g$. In case of movement, the external force $a^{(l)}$ interacts additively \cite{kok18}. Accelerometers are suitable for long-term measurements due to absence of drifts and constant positions of earth's gravity centre \cite{car16}. However, noise behaviour is evident. The lack of information about yaw $\theta$ allows correct tilts only for roll $\psi$ and pitch $\phi$ \cite{vec22}.

The magnetometer enables the determination of $\theta$ since the actual direction of $\phi,\theta,\psi$ depends on latitude and longitude \cite{vec22}. They refer to an Earth-Fixed Coordinate System (ECEF). This transformation from inertial to earth-fixed coordinate system is described as a rotation since a common reference is used. 

\textbf{Magnetic Inertial Navigation}
A further method for inertial navigation is the position determination by means of magnetic sensors. 

Shi et al. \cite{shi21} examined the navigated indoor positioning using optimisation algorithms for magnetic reference maps. They demonstrated in their motion experiment that positional accuracy and matching with inertial navigation devices is significantly improved through magnetic references.

Kok et al. \cite{kok18} demonstrated empirically that magnetic field maps achieve efficient position estimates. They identified the necessary proximity between sensor and magnetic field generating coils for radial positions as well as altitude error reduction and concluded that further magnetic disturbance decreases the information content of measurements.

Caruso et al. \cite{car16} showed that fused estimation of magnetometer arrays and VINS are able to reconstruct outdoor trajectories where Magneto-Inertial Dead-Reckoning (MI-DR) techniques fail due to gradient leaks. However, magnetic estimation techniques and leakage of suppressed magnetic information need to be improved. They concluded that magnetic navigation could expand the application range in unfavourable environments and reduce power consumptions.

\begin{figure*}[ht!]{}
    \centering
    \hspace*{0cm} \includegraphics[width=0.9\textwidth]{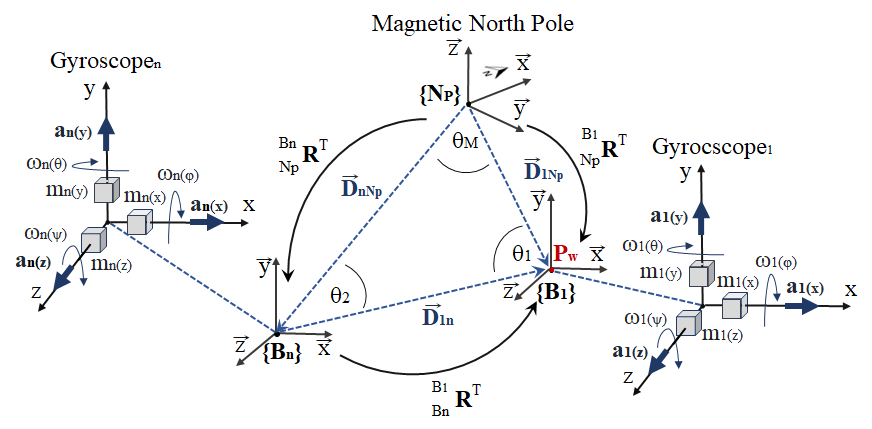}
    \caption{\textbf{Conceptual Representation of Global Sensor Matching}: Description of the geometric relationships between the sensor points $\{ B_\mathrm{1}\}$ ,$\{B_\mathrm{n}\}$ and common origin $\{N_\mathrm{p}\}$ in the WCS.}
    \label{fig:ConceptDynamicSensorMatching}
\end{figure*}

\textbf{VINS} Initial orientations of navigated devices can be estimated with IMUs. The combined merging of camera and IMU data allows extensive image information for efficient position solutions \cite{yan19}. VINS use points and lines with online spatial and temporal calibrations. Additional feature observations from different keyframes enable a reduction of trajectory sensor drifts \cite{hua19}. 
\newpage
Huang \cite{hua19} investigated short-term compatibility for 3D motion tracking by comparing VIO and SLAM for local navigation. VINS is not suitable for long-term, large-scale, safety-critical deployments and under difficult conditions. This was mainly due to poor illumination and movement. Geometric features such as points, lines and areas, as used in current VINS for localisation, are unsuitable for semantic localisation and mapping. The real-time implementation of VINS continues to be challenging, despite initial efforts. Auxiliary sensors for specific environments and movements, such as sonar or LiDAR, enable better detection of dynamic movements. Huang asserts that simple integration of high-frequency IMU measurements is unreliable for long-term navigation, due to noise and distortion behaviour.

Tardif et al. \cite{tar10} demonstrated a robust image-based inertial navigation system for rural and urban environments based on VINS. Experimentally, the prototype showed low deviations at a maximum speed of $70 km/h$.

In \cite{vu12}, experimental results have shown that the combination of DGPS and a single visual feature measurement at $1 Hz$ is sufficient to achieve $1 m$ positional accuracy.
Vu et al. used visual features like traffic lights to generate a better positional and situational awareness in their tightly coupled real-time vision and DGPS-based INS. 

\section{Global Sensor Matching}
This section describes the underlying concept for global matching of multiple sensors in a new WCS. We define the first local sensor as the new origin of this WCS and transform the initialised data of all remaining sensors once to the first sensor point $\{B_\mathrm{1}\}$. This enables future extensible matchings with VINS or feature detection and additional inclusions of local fused sensor information.

To represent movements, it is necessary to transform intrinsic coordinates of depth and position sensors into a new WCS. We observe the IMU integrated gyroscope and magnetometer of two different stereo cameras ($1$ and $n$). A common reference coordinate allows the calculation of positional IMU relationships. The defined points $\{B_\mathrm{1}\}$ and $\{B_\mathrm{n}\}$ of the magnetometers refer to the magnetic North Pole as common origin $\{N_\mathrm{p}\}$ (Fig.~\ref{fig:ConceptDynamicSensorMatching}).

From this consideration we can form the mathematical relationships between $\{N_\mathrm{p}\}$, $\{B_\mathrm{1}\}$ and $\{B_\mathrm{n}\}$. The point $P_\mathrm{w}$ marks the new origin of our common WCS ($P_\mathrm{w} \hat{=} \{ B_\mathrm{1}\}$). We define the scalar values $\vec{D}_\mathrm{nNp}$, $\vec{D}_\mathrm{1n}$ and $\vec{D}_\mathrm{1Np}$ between $\{N_\mathrm{p}\}$, $\{B_\mathrm{1}\}$, $\{B_\mathrm{n}\}$ (Fig.~\ref{fig:ConceptDynamicSensorMatching}) and denote the rotational coordinate transformations as $^{B_\mathrm{n}}_{N_\mathrm{p}}R^{T}$, $^{B_\mathrm{1}}_{B_\mathrm{n}}R^{T}$ and $ ^{B_\mathrm{1}}_{N_\mathrm{p}}R^{T}$. 

The following relationships apply to the transformation of $\{B_\mathrm{n}\}$, $\{B_\mathrm{1}\}$ and $\{N_\mathrm{p}\}$: 
\begin{equation}\label{eq:E1}
      B_\mathrm{n} = \hspace*{0.1cm} ^{B_\mathrm{n}}_{N_\mathrm{p}}R^{T} \cdot ({N}_\mathrm{p} - \vec{D}_\mathrm{nNp}) 
\end{equation}
\begin{equation}\label{eq:E2}
      B_\mathrm{1} = \hspace*{0.1cm} ^{B_\mathrm{1}}_{B_\mathrm{n}}R^{T} \cdot ({B}_\mathrm{n} - \vec{D}_\mathrm{1n}) \hspace*{0.2cm}
\end{equation}
\vspace*{-0.15cm}
\begin{equation}\label{eq:E3}
      B_\mathrm{1} = \hspace*{0.1cm} ^{B_\mathrm{1}}_{N_\mathrm{p}}R^{T} \cdot ({N}_\mathrm{p} - \vec{D}_\mathrm{1Np})
\end{equation}
\begin{equation}\label{eq:D1np}
      \vec{D}_\mathrm{1N_\mathrm{p}} = \vec{D}_\mathrm{nN_\mathrm{p}} + \vec{D}_\mathrm{1n}
\end{equation}

By transforming Eq.~\ref{eq:E1} and Eq.~\ref{eq:E3} according to $\{N_\mathrm{p}\}$, we obtain the distance $\vec{D}_\mathrm{1n}$ between the position of the magnetic sensors $\{B_\mathrm{1}\}$ and $\{B_\mathrm{n}\}$.
\begin{equation}\label{eq:EqualEquationD1n}
    \dfrac{B_\mathrm{n}}{^{B_\mathrm{n}}_{N_\mathrm{p}}R^{T}} + \vec{D}_\mathrm{nNp}= \dfrac{B_\mathrm{1}}{^{B_\mathrm{1}}_{N_\mathrm{p}}R^{T}} +  (\vec{D}_\mathrm{nN_\mathrm{p}} + \vec{D}_\mathrm{1n})
\end{equation}
\begin{equation}\label{eq:EquationD1n}
    \hspace*{0.3cm} \vec{D}_\mathrm{1n} = \dfrac{B_\mathrm{n}}{^{B_\mathrm{n}}_{N_\mathrm{p}}R^{T}} - \dfrac{B_\mathrm{1}}{ ^{B_\mathrm{1}}_{N_\mathrm{p}}R^{T}}
\end{equation}
Eq.~\ref{eq:EquationD1n} shows the relationship between $\{B_\mathrm{n}\}$ and $\{B_\mathrm{1}\}$.
The initial position of $\{B_\mathrm{1}\}$ is only influenced by the rotation ratio between $\{B_\mathrm{n}\}$ and $\{B_\mathrm{1}\}$ in case of $|D_\mathrm{1n}| = 0$. The related distance dependencies of $\{B_\mathrm{1}\}$,$\{B_\mathrm{n}\}$ to $\{N_\mathrm{p}\}$ can be determined by substituted vector length $D_\mathrm{1n}$ from Eq.~\ref{eq:D1np} to Eq.~\ref{eq:EquationD1n}.
\begin{equation}\label{eq:DistanceD1NpDnNp}
   \vec{D}_\mathrm{1N_\mathrm{p}} - \vec{D}_\mathrm{nN_\mathrm{p}}  = \dfrac{B_\mathrm{n}}{^{B_\mathrm{n}}_{N_\mathrm{p}}R^{T}} - \dfrac{B_\mathrm{1}}{ ^{B_\mathrm{1}}_{N_\mathrm{p}}R^{T}}
\end{equation}
The new coordinate system at origin $\{B_\mathrm{1}\}$ can be calculated using Eq.~\ref{eq:E2}. Therefore we substitute Eq.~\ref{eq:EquationD1n} into Eq.~\ref{eq:E2}. 
\begin{equation}\label{eq:EquationB1}
    B_\mathrm{1} =  ^{B_\mathrm{1}}_{B_\mathrm{n}}R^{T} \cdot \biggr( {B}_\mathrm{n} - \biggr(\dfrac{B_\mathrm{n}}{^{B_\mathrm{n}}_{N_\mathrm{p}}R^{T}} - \dfrac{B_\mathrm{1}}{ ^{B_\mathrm{1}}_{N_\mathrm{p}}R^{T}}\biggr) \biggr)
\end{equation}
We receive the following result solving Eq.~\ref{eq:EquationB1}:
\begin{equation}\label{eq:ResultingEquationB1}
    B_\mathrm{1} = {B}_\mathrm{n} \cdot \hspace*{0.1cm} ^{B_\mathrm{1}}_{B_\mathrm{n}}R^{T} \cdot \dfrac{\biggr( 1 - \dfrac{1}{^{B_\mathrm{n}}_{N_\mathrm{p}}R^{T}} \biggr)}{\biggr( 1 - \dfrac{^{B_\mathrm{1}}_{B_\mathrm{n}}R^{T}}{ ^{B_\mathrm{1}}_{N_\mathrm{p}}R^{T}}  \biggr)} 
\end{equation}
We combine the rotational transformations of Eq.~\ref{eq:ResultingEquationB1} to $R_\mathrm{B_\mathrm{1}}$.
\begin{equation}\label{eq:RB1}
    B_\mathrm{1} = {B}_\mathrm{n} \cdot R_\mathrm{B_\mathrm{1}} 
\end{equation}
The trajectory position information of $R,T$ can be calculated by integrated velocity over a discrete time period. We can determine the measured acceleration $a$ and angular velocity $\omega$ in the body frame oriented IMU. The associated determination of Euler angles in the direction cosine matrix $C$ allows the velocity transformation from body$^{b}$ to inertial$^{i}$ frame \cite{vec22} (Eq.~\ref{eq:DCM}). 
\begin{equation}\label{eq:DCM}
    C^{b} \approx  \left[
                \begin{array}{c c c}
                    1 & \psi & -\theta\\
                    -\psi & 1 & \phi\\ 
                    \theta & -\phi & 1\\  
                \end{array}
                \right]
\end{equation}
\begin{equation}\label{eq:InertialVelocity}
    \Delta v^{i} = C^{b} \int_{t_\mathrm{n}}^{t_\mathrm{n}+\Delta} a^{b}_\mathrm{lin} \,dt
\end{equation}
The accelerometer integrated seismic mass affects motion-dependent special force measurements of coriolis $a_\mathrm{cl}^{b}$, centrifugal $a_\mathrm{cf}^{b}$ and gravitational $a_\mathrm{g}^{b}$ acceleration (Eq.~\ref{eq:VectorResultingAcceleration}) \cite{cla15,kok18,vec22}.
\begin{equation}\label{eq:ResultingAcceleration}
    a_\mathrm{cor}^{i} = a_\mathrm{lin}^{b} + a_\mathrm{cf}^{b} + a_\mathrm{cl} - a_\mathrm{g}^{b} \hspace*{0.4cm} \{ \hspace*{0.2cm} a_\mathrm{cl}^{b} \ll a_\mathrm{cf}^{b}
\end{equation}
\begin{equation}\label{eq:VectorResultingAcceleration}
   a_\mathrm{cor}^{i} =  a_\mathrm{lin}^{b} + \omega \times v^{b} - R^{b} \cdot a_\mathrm{g}^{b} \hspace{1.65cm}
\end{equation}
\begin{figure*}[!ht]{}
    \centering
    \includegraphics[width=0.8\textwidth]{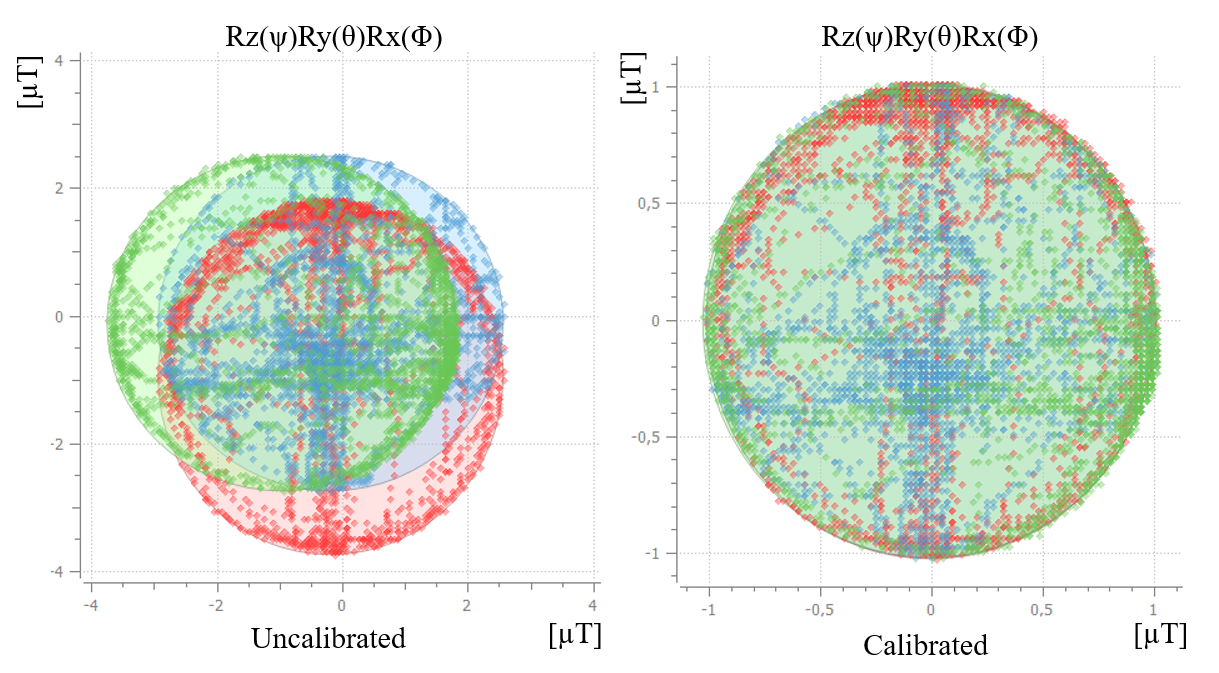}
    \caption{\textbf{Calibration of the Magnetic Field Sensor}: The magnetometer is integrated in the chassis of the ZED 2 from Stereolabs. Each colour represents a direction of rotation (red: $R_\mathrm{\phi}$, green: $R_\mathrm{\theta}$, blue: $R_\mathrm{\psi}$).}
    \label{fig:CalibrationPitchRollYaw}
\end{figure*}
Assuming that the navigation frame is fixed on earth's frame position, we can derive a relation for the corrected velocity (Eq.~\ref{eq:VectorResultingVelocity}) \cite{kok18,vec22} . 
\begin{equation}\label{eq:VectorResultingVelocity}
    \Delta v_\mathrm{cor}=  \Delta v^{I} + \Delta t \bigl( a_\mathrm{g}^{b} - 2 \omega_\mathrm{\oplus} \times \int_{t_\mathrm{0}}^{t_\mathrm{n}}  \hspace{-0.1cm} a_\mathrm{}^{i} \hspace{0.1cm} \,dt \bigl)
\end{equation}
$\omega_\mathrm{\oplus}$ defines the earth's angular rate \cite{cla15}. The earth fully rotates every 23.9345 hours with an approximated rate of $\omega_\mathrm{\oplus} = 7.29 \cdot 10^{-5}$ $rad/s$ relative to the stars \cite{kok18}. The position $s$ can be calculated from the relationships of Eq.~\ref{eq:VectorResultingVelocity}. 
\begin{equation}\label{eq:VelocityResultV1}
    \Delta v_\mathrm{k+1}= \int_{t_\mathrm{0}}^{t_\mathrm{n}}  \hspace{-0.1cm} a_\mathrm{cor} \hspace{0.1cm} \,dt + \Delta v_\mathrm{cor}
\end{equation}
\begin{equation}\label{eq:PositionResultV1}
    \Delta s_\mathrm{k+1}= \bigl(\iint_{t_\mathrm{0}}^{t_\mathrm{n}} a_\mathrm{cor}^{i} \,dt \bigl)+ \Delta t \bigl(\int_{t_\mathrm{n}}^{(t_\mathrm{n}+\Delta)} \hspace*{-0.4cm} a_\mathrm{lin} \,dt\bigl) + \dfrac{\Delta t}{2} \Delta v_\mathrm{cor}
\end{equation}

\section{Methodology of Initial Space Movement}
\begin{figure}[ht!]
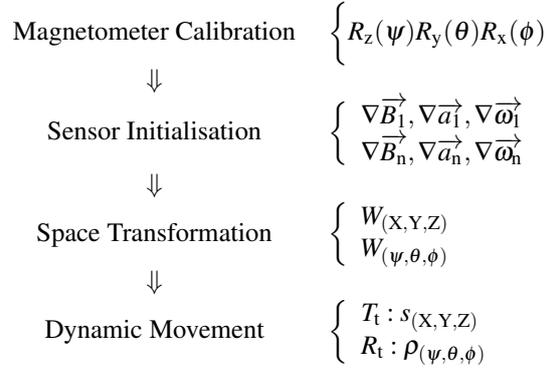

    \centering
    \caption{\textbf{Pipeline of Space Determination:} The different steps of movement transformations in a WCS.}
    \label{fig:PipelinePositionDetermination}
\end{figure}

Our methodological approach as shown in Fig.~\ref{fig:PipelinePositionDetermination} can be classified into the steps of calibration, initialisation, transformation and locomotion. 

Since we define a magnetically referenced WCS at timestep $t_\mathrm{0}$, small positional fluctuations are inevitable in the presence of external influences. A general calibration is carried out for this reason. The sensor initialisation allows the determination of a specified transfer function for unit transformation $\nabla B \rightarrow \nabla s$. The space transformation converts the intrinsic sensor coordinates into the magnetically defined WCS. Using the motion equations in case of dynamic movement allows the space positional translation and rotation.

\begin{table}[ht!]
\begin{tabular}{c l}
Magnetometer Calibration &  $ \biggl\{ R_\mathrm{z}(\psi) R_\mathrm{y}(\theta) R_\mathrm{x}(\phi)$ \\
$\Downarrow$ & \\
Sensor Initialisation &  $\biggl\{ \begin{array}{l}   
\nabla \overrightarrow{B_\mathrm{1}},
\nabla \overrightarrow{a_\mathrm{1}},
\nabla \overrightarrow{\omega_\mathrm{1}} \\
\nabla \overrightarrow{B_\mathrm{n}},
\nabla \overrightarrow{a_\mathrm{n}},
\nabla \overrightarrow{\omega_\mathrm{n}} \\
\end{array}$  \\
$\Downarrow$ & \\
Space Transformation &  $\biggl\{ \begin{array}{l}   
W_\mathrm{(X,Y,Z)} \\
W_\mathrm{(\psi,\theta, \phi)}
\end{array}$  \\
$\Downarrow$ & \\
Dynamic Movement & $\biggl\{ \begin{array}{l}   
T_\mathrm{t}: s_\mathrm{(X,Y,Z)} \\
R_\mathrm{t}: \rho_\mathrm{(\psi,\theta, \phi)}
\end{array} $  \\
\end{tabular}
\vspace*{-0.7cm}
\end{table}

The accuracy of magnetic field sensors is affected by external temperature as well as ferro-, para- and diamagnetic influences \cite{cla15,ren10,vec22}. \newpage The homogeneity of earth's magnetic field is determined by local anomalies, dipole and external fields. The earth field's detectability depends on the sensitivity of the magnetic field sensor \cite{cla15}. Since these effects can occur in- and outdoors, magnetometers must be calibrated. Fig.~\ref{fig:CalibrationPitchRollYaw} demonstrates the difference between a calibrated and uncalibrated sensor. Under ideal conditions, the measured points (X,Y,Z) are located at the centre $\{$0,0,0$\}$ of overlayed perfect spheres (RGB) since their position changes in all spatial directions. The non-calibrated state on the left side of Fig.~\ref{fig:CalibrationPitchRollYaw} shows the inconsistent relationship between the axes (red: $R_\mathrm{\phi}$, green: $R_\mathrm{\theta}$, blue: $R_\mathrm{\psi}$). This shift condition can be corrected by the Hard and Soft Iron calibration \cite{ren10,vec22}. 

The magnetic disturbances and error sources can be corrected mathematically. Therefore, we consider the magnetometer $B_\mathrm{c}$ with the Eq.~\ref{eq:DCM} based transformation matrix $C$ and temperature dependent bias $b_\mathrm{H}$ to the vector of non-calibrated magnetic field data $\tilde{B}$ \cite{ren10,vec22}.
\begin{equation}\label{eq:CalibrationMagnetometer}
\left[
\begin{array}{l}
    B_\mathrm{cx} \\
    B_\mathrm{cy} \\ 
    B_\mathrm{cz} \\  
\end{array}
\right] = 
\left[
\begin{array}{c c c}
    C_\mathrm{00} & C_\mathrm{01} & C_\mathrm{02}\\
    C_\mathrm{10} & C_\mathrm{11} & C_\mathrm{12}\\ 
    C_\mathrm{20} & C_\mathrm{21} & C_\mathrm{22}\\  
\end{array}
\right]
\left[
\begin{array}{c}
    \tilde{B}_\mathrm{x} - b_\mathrm{H_\mathrm{0}} \\
    \tilde{B}_\mathrm{y} - b_\mathrm{H_\mathrm{1}} \\ 
    \tilde{B}_\mathrm{z} - b_\mathrm{H_\mathrm{2}} \\  
\end{array}
\right]
\end{equation}
Rotating the magnetometer around the gravity vector for several 360$^{\circ}$ cycles allows calibration. The sensor should inclinate between 5$^{\circ}$ and 10$^{\circ}$ \cite{vec22}. A successful calibration contains the measuring points of all rotation directions in a circle (Right diagram - Fig.~\ref{fig:CalibrationPitchRollYaw}).

\begin{figure}[ht!]
    \centering
    \hspace*{-0.2cm} \includegraphics[scale=0.38]{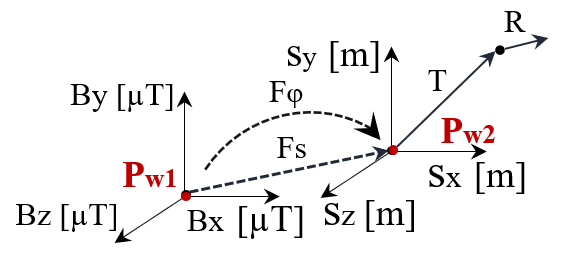}
    \caption{\textbf{Unit Transformation $[\mu T] \rightarrow[m]$}: Transformation of the magnetic field density B[$\mu T$] into the distance $s$[$m$] ($F_\mathrm{s} \cdot \nabla B[\mu T] \rightarrow  \nabla s[m] $). $F_\mathrm{s}$ and $F_\mathrm{\varphi}$ describes the transfer function from $[\mu T]$ to $[m]$.}
    \label{fig:koordinatentransformation}
\end{figure}

The next step refers to the initial position of the different sensor parameters. $P_\mathrm{w1}$ (left coordinate system in Fig.~\ref{fig:koordinatentransformation}) is at the origin of our magnetic WCS $\{B_\mathrm{1}\}$ at the timestep $t_\mathrm{0}$.  
The susceptibility to external influences makes the magnetic measurement data unsuitable for permanent dynamic position determination. For this reason, we transfer the magnetic unit to a metre unit (right coordinate system in Fig.~\ref{fig:koordinatentransformation}) and initialise it with the measurement data of the gyroscope as well as the accelerometer. The initialisation and measurement fusion of all available sensors in the transformed metric system helps to reduce measurement fluctuations and influences of a single sensor.
\vspace*{-0.4cm}

The origin $P_\mathrm{w1}$, related to $\{B_\mathrm{1}\}$, can be calculated with Eq.~\ref{eq:ResultingEquationB1}. The IMU integrated magnetometer and gyroscope have a sensor specific step size during the spatial movement $(R|T)$. Gradient formation enables the conversion of the step size to a unit of measurement. The transfer functions of translation $F_\mathrm{s}$ and rotation $F_\mathrm{\varphi}$ describe the unit transformation from magnetic $\nabla B$ to distance $\nabla s$ at $t_\mathrm{}$:
\begin{equation}\label{eq:Fs}
   \hspace*{0.35cm}  F_\mathrm{s} = \dfrac{\nabla s_\mathrm{}}{\nabla B_\mathrm{}} \hspace{0.2cm} \dfrac{[m]}{[\mu T]}  \hspace{0.4cm} \biggl\{
    \begin{array}{l}
        \hspace{0.1cm} t_\mathrm{} > 0 \\
        \hspace{0.0cm} B_\mathrm{} > 0
    \end{array}
\end{equation}
\begin{equation}\label{eq:Fphi}
    \hspace*{-0.2cm} F_\mathrm{\varphi} = \hspace{0.1cm} ^{P_\mathrm{w2}}_{P_\mathrm{w1}}R^{T} \hspace{0.2cm} [^{\circ}] \hspace*{0.52cm} \biggl\{ \hspace{0.1cm} t_\mathrm{} > 0
\end{equation}
Since the distances between the components of gyroscope, acceleration and magnetic field sensor are sufficiently small, the coordinate systems converge ($F_\mathrm{s} \cdot \Delta B \rightarrow \Delta s$). The initial translation and rotation values $\nabla s$ and $\nabla \varphi$ can be used approximately as starting values due to converging systems. 
\vspace*{-0.2cm}
\begin{equation}
  R \rightarrow \varphi [^{\circ}]:  \hspace{0.2cm}  \Delta \varphi_\mathrm{(t_\mathrm{})} =    F_\mathrm{\varphi} + \int  \hspace*{-0.0cm} \omega_\mathrm{} \,dt \hspace*{0.784cm}
\end{equation}
\begin{equation}
  \hspace*{0.2cm} T \rightarrow s [m]:  \hspace{0.2cm}  \Delta s_\mathrm{(t_\mathrm{})} =  F_\mathrm{s} \cdot \nabla B_\mathrm{}  + \iint_{}^{}   \hspace*{-0.0cm} a_\mathrm{}^{} \,dt
\end{equation}
The transfer functions $F_\mathrm{s}$ and $F_\mathrm{\varphi}$ can be determined for each sensor (Eq.~\ref{eq:Fs} and Eq.~\ref{eq:Fphi}). This allows the sensor unit-specific transmission of measured values in the WCS. The direct sensor relations to each other can be determined. Fluctuations, noise or susceptibilities due to external influences are evident. By comparing the position data between two defined coordinate systems $P_\mathrm{w1}$ and $P_\mathrm{w2}$, magnetic interference or sensor-specific drift deviations can be corrected. 

The analysing of locomotive position sensors is an important part for the validation of dynamic matching. Therefore, we focused on the differences between static and dynamic states of multiple matched sensors and the extent of environmental influences in our methodological considerations.

\begin{figure}[ht!]
    \centering
    \includegraphics[scale=0.30]{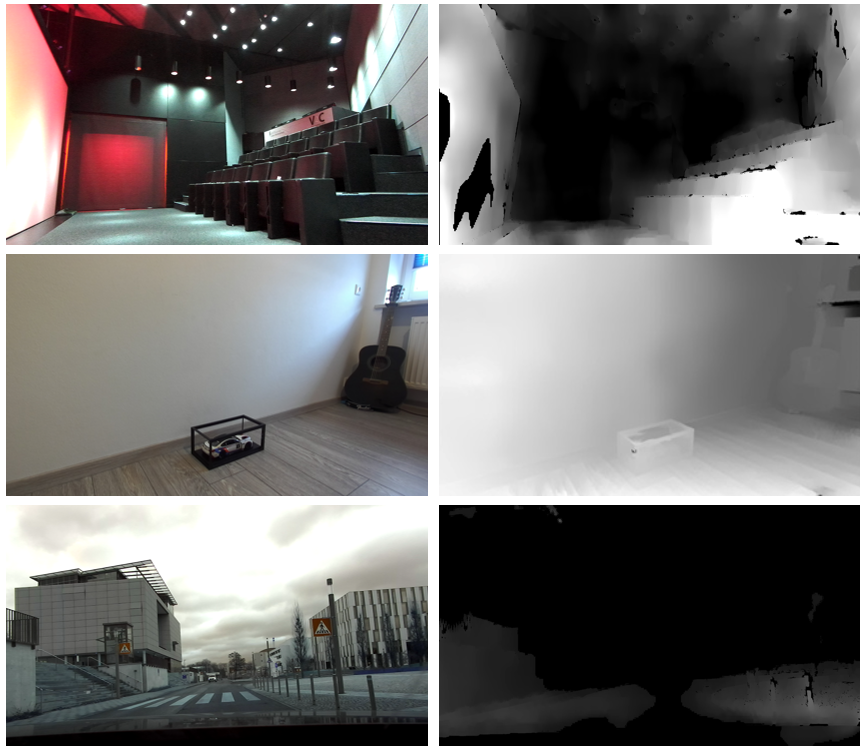}
    \caption{\textbf{Experimental Setup:} Selected measurement locations (left side) and their associated depth images (right side). Top: Indoor space of a Lab; Middle: Model vehicle in a residential building; Bottom: Outside in a moving vehicle}
    \label{fig:ExperimentalSetup}
\end{figure}

We select different experimental locations (Fig.~\ref{fig:ExperimentalSetup}) for our validation and gather criteria (Tab.~\ref{fig:CriteriaMeasurement}) such as ferromagnetic influenced building materials, lighting and colour conditions, textures as well as velocities from the challenges of previous works. 

\vspace*{-0.2cm}
\begin{center}
\begin{table}[ht!]
\begin{tabular}{r c c c}
\hline
& \textbf{Lab} & \textbf{Model} & \textbf{Vehicle} \\
\hline
& & & \vspace*{-0.3cm} \\
Velocity [$km/h$] & 0 < 3.6 & 0 & 0 < 80 \\
Location size [m] & 30 < & 1 < & < 2000 \\
Magnetic influences & Indoor & Indoor & Outdoor \\
Texture variations & Low & Low & High \\
Light & Low & Average & High \\
Colour Contrasts & Low & Average & High \\
& & & \vspace*{-0.3cm} \\
\hline
\end{tabular}
\caption{\textbf{Criteria of Measurement:} Description of the selection criteria for different experimental locations.}
\label{fig:CriteriaMeasurement}
\end{table}
\end{center}
\vspace*{-0.5cm}
\begin{table*}[!htbp]
\centering
\begin{tabular*}{0.98\textwidth}{r c c c c c}
\hline
& $\sigma_\mathrm{nc}$ [$\mu T$] & $\sigma_\mathrm{c}$ [$\mu T$] & $\epsilon_\mathrm{c/nc}$ [$\%$] & Uncalibrated & Calibrated \\
\hline
& & & & &\vspace*{-0.3cm} \\
\hspace*{0.0cm} \parbox{1.0cm}{\hspace*{-0.45cm} \textbf{Model:}} & \parbox{3cm}{$3.00 \pm 12.86 \cdot 10^{-3} $} & \parbox{3cm}{$2.05 \pm 23.00  \cdot 10^{-3}$} & \parbox{1cm}{$31.67$} & \parbox{1cm}{\includegraphics[scale=0.15]{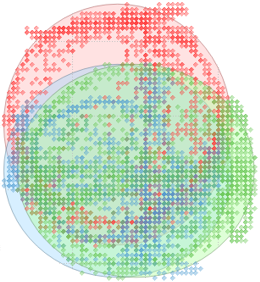}} &  \parbox{1cm}{\includegraphics[scale=0.13]{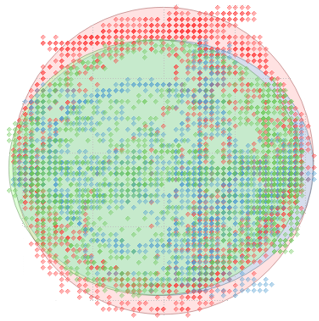}} \\
\hspace*{0.0cm} \parbox{1cm}{\textbf{Lab:}} & \parbox{3cm}{$0.99 \pm 11.10 \cdot 10^{-3} $} & \parbox{3cm}{$0.97 \pm 8.09 \cdot 10^{-3}$} & \parbox{1cm}{$2.51$} & \parbox{1cm}{\includegraphics[scale=0.13]{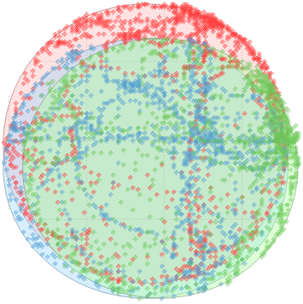}} &  \parbox{1cm}{\includegraphics[scale=0.12]{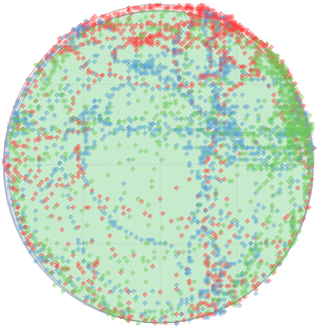}} \\
\hspace*{0.0cm} \parbox{2.4cm}{\textbf{ICEV Centre:}} & \parbox{3cm}{$1.29 \pm 0.00 \cdot 10^{-3} $} & \parbox{3cm}{$1.28 \pm 10.83 \cdot 10^{-3}$} & \parbox{1cm}{$0.35$} & \parbox{1cm}{\includegraphics[scale=0.15]{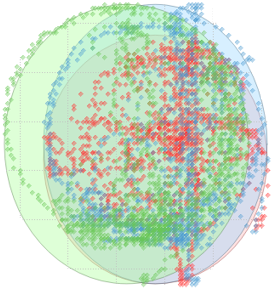}} &  \parbox{1cm}{\includegraphics[scale=0.15]{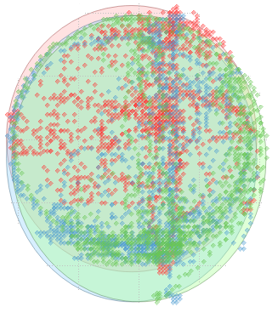}} \\
\hspace*{0.0cm} \parbox{2.5cm}{\hspace*{0.2cm} \textbf{ICEV Front:}} & \parbox{3cm}{$6.72 \pm 3.11 \cdot 10^{-3} $} & \parbox{3cm}{$1.52 \pm 11.48 \cdot 10^{-3}$} & \parbox{1cm}{$77.50$} & \parbox{1cm}{\includegraphics[scale=0.14]{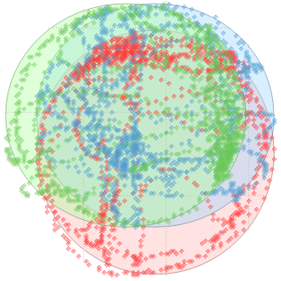}} &  \parbox{1cm}{\includegraphics[scale=0.13]{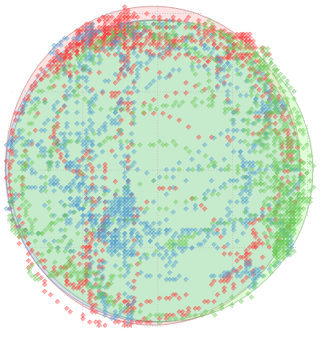}} \\
\hspace*{0.0cm} \parbox{2.0cm}{\textbf{EV Centre:}} & \parbox{3cm}{$2.56 \pm 0.01 \cdot 10^{-3} $} & \parbox{3cm}{$0.96 \pm 15.39 \cdot 10^{-3}$} & \parbox{1cm}{$62.53$} & \parbox{1cm}{\includegraphics[scale=0.14]{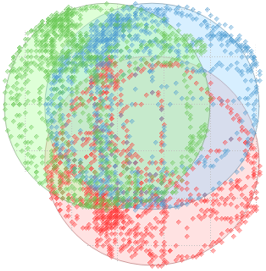}} &  \parbox{1cm}{\includegraphics[scale=0.13]{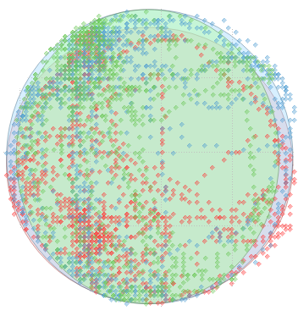}} \\
\hspace*{0.0cm} \parbox{2.1cm}{\hspace*{0.2cm} \textbf{EV Front:}} & \parbox{3cm}{$1.89 \pm 2.22 \cdot 10^{-16} $} & \parbox{3cm}{$1.15 \pm 10.16 \cdot 10^{-3}$} & \parbox{1cm}{$39.27$} & \parbox{1cm}{\includegraphics[scale=0.14]{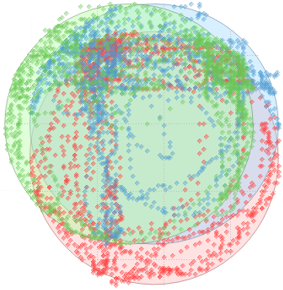}} &  \parbox{1cm}{\includegraphics[scale=0.13]{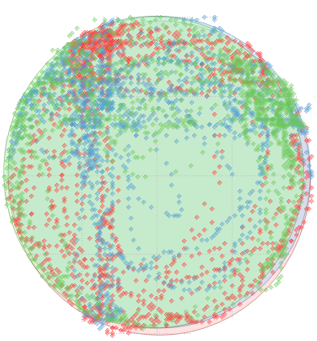}} \\
& & & & &\vspace*{-0.3cm} \\
\hline
\end{tabular*}
\caption{\textbf{Stability of the Magnetometer:} $\sigma_\mathrm{c}$ and $\sigma_\mathrm{nc}$ ($\sigma_\mathrm{c}$ :calibrated, $\sigma_\mathrm{nc}$: non-calibrated) describe the deviation of magnetic values during measured time $t_\mathrm{c}$ and $t_\mathrm{nc}$ ($t_\mathrm{c}$: calibrated, $t_\mathrm{nc}$: non-calibrated). The calibrated and uncalibrated states of the magnetometer are represents roll: red, pitch: green and yaw: blue. }
\label{fig:MagnetometerEvaluation}
\end{table*}
\section{Evaluation}
We investigate the static and dynamic behaviour of ZED 2 integrated sensors for the evaluation of geomagnetic inertial navigation and positional sensor stability. 
Our computing hardware has an integrated Intel Core i5 processor (@2.5GHz), 64GB RAM, external GPU1 Nvidia GeForce GTX 1050 Ti and onboard GPU0 of Intel HD Graphics 630.

\textbf{Static Sensor State} The stability behaviour is evaluated through revision of ZED 2 integrated magnetometer, accelerometer and gyroscope in stationary case. We consider local measurement fluctuations (Fig.~\ref{fig:ExperimentalSetup}) in calibrated and uncalibrated state to explore the suitability of geomagnetic initialisation at different times. In addition to different indoor spaces, we examine sensor behaviours in Internal-Combustion-Engine Vehicles (ICEV) and Electric Vehicle (EV). We investigate static influences on magnetometers caused by Faraday cage, electric motors, battery, active sensors, embedded graphic and combination instruments. Our static results are shown in Tab.~\ref{fig:MagnetometerEvaluation}.

\textbf{Static Evaluation} The location-dependent comparison in our evaluation show necessity of magnetometer calibration and accelerometer compension in static case. The magnetometer reveal local susceptibilities. Electromagnetic fields from devices act on the sensor at spatial locations and within the vehicles. Previous sensor positions or micro movements deviate from actual calibrated conditions between 31.67$\%$ and 77.5$\%$. We notice in the vehicles (ICEV and EV) that magnetometers could not be calibrated at any local position due to possible magnetic fields. A strong vulnerability occurs in front of the position area where fluctuations are measurable despite calibrated states. The centre vehicle area of ICEV prove as well suitable for calibration position. We suspect that engine compartments and embedded graphic and combination instruments induce fields. The Faraday cage exhibit no direct effects in our measurements. Contrary to the ICEV, we are able to achieve stable measurement results at several static positions in the EV. Both vehicles show the consistently best results at the calibrated position. A sensor shift from the calibrated location cause a direct deterioration of position accuracy.
We conclude that the magnetometer is suitable for one-time initialisation. The ideal moment is immediately after calibration. Compressing accelerometers drift and gravity at the time of magnetic calibration allows efficient positional accuracy. 

\textbf{Dynamic Sensor State} The accuracy and transferability of multiple sensor trajectory representations is directly related to the sensor behaviour comprising stability, resolution, accuracy and speed response. Sensory noises and limitations correlate with the necessary tolerance band of trajectory positions. However, keeping within the tolerance band is an important criterion for the successful implementation of our methodology. Therefore, we compare and analyse the dynamic behaviour with the resting IMU state.

\begin{figure}[ht!]
    \centering
    \includegraphics[scale=0.35]{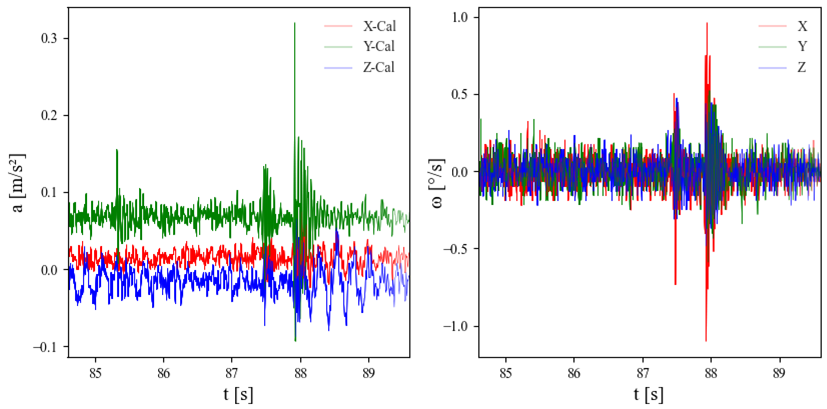}
    \caption{\textbf{Sensory Ground Truth:} The left diagram shows the acceleration. The right diagram represents the angular velocity.}
    \label{fig:NoiseSensors}
\end{figure}

\begin{figure}[ht!]
    \centering
    \includegraphics[scale=0.33]{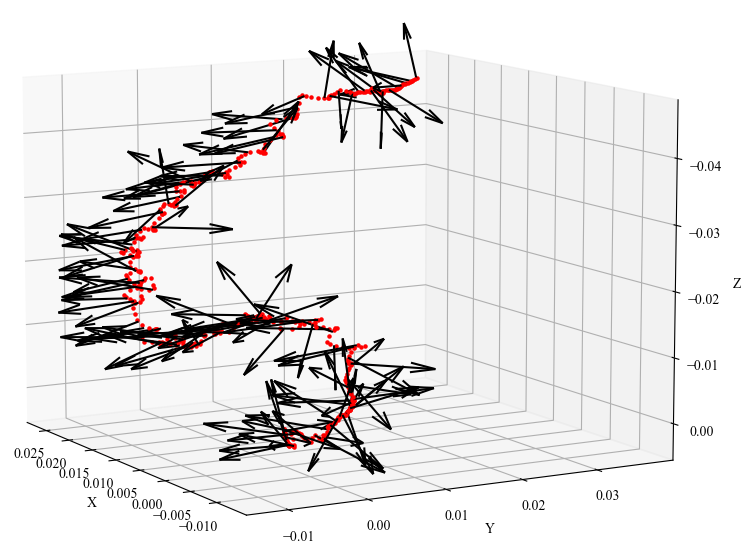}
    \caption{\textbf{Trajectory Representation of Climb Stairs:} Black arrows show the sensor orientation. Red dots represent the calculated position.}
    \label{fig:SensorTrajecotries}
\end{figure}

\textbf{Dynamic Evaluation} In resting state, the linear acceleration (Eq.~\ref{eq:Accelerometer}) and angular velocity (Eq.~\ref{eq:GyroscopeVelocity}) exhibit noise and offset behaviours in all directions (Fig.~\ref{fig:NoiseSensors}).
\textbf{}
As a result, the double integrated acceleration after time leads to major position deviations. Adjusting the offset reduces the deviation error but does not eliminate it. By implementing a Kalman and low-pass filter, we are able to improve the noise and following position behaviour. Especially, Kalman filter is suitable for the fusion of several sensor signals in a dynamic system, which means that we can compensate the accelerometer signal error with the magnetometer. 
The section of human positional trajectory representation (Fig.~\ref{fig:SensorTrajecotries}) shows inaccuracies of sensor orientation (black arrows). 

We expect that the human sinusoidal oscillation leads to a harmonic behaviour of the sensor alignment. However, our results show a non-harmonic progression. The orientation inaccuracies result from noise behaviour and sensitivity of the measured angular velocity. The use of Kalman and low-pass filters may also increase the orientation quality.
\section{Conclusion and Future Work}
In this paper we present a concept for global and dynamic matching of multiple depth cameras using dynamic sensor data. Our motivation is a stable and verifiable inertial navigation system for in- and outdoor depth sensing. Limiting ranges and external influences such as contrast, textures, temperature, magnetic interferences and insulating materials affect the accuracy of conventional methods like GPS or image initialisation. Inaccurate position data decrease the quality of matched point clouds due to visible scattering, distortion, noise and offsets effects. Based on these challenges, we propose a concept to transfer multi-sensor data into a magnetically referenced WCS: the geomagnetic field. Global depth sensor matching allows the environmental reconstruction of individual geographic positions, while alternative navigation systems are insufficient. Geomagnetic sensor matching can be used wherever a stable external magnetic field is measured.

The positional matching of dynamic depth sensors is a promising technique for the expanding and optimising of 3D reconstructions. The suitability of global adjustment for in- and outdoor applications is based on the measurability of earth's magnetic field. A WCS can be generated by magnetic field sensors. Referencing the geomagnetic North Pole allows a direct proportionality of the magnetic field sensors. Coordinate transformations between the magnetometer and IMU can compensate the magnetic susceptibility, drifts and noise effects. The sensors show fluctuations during our in- and outdoor measurements. Combining multiple sensors reduce the position error and following offsets in merged point clouds. Calibration of magnetic field sensors increases stability of measurements despite magnetic interference sources. Future approaches can implement Kalman or low-pass filters to decrease integration-related position and orientation deviations. 

The measurement radius will be extended for future evaluations of external influences and functionalities of geomagnetic matching. We will expand our data sets with different geographical locations, higher movement speeds, long-term measurements and in- and outdoor combination. Extended data sets enable us to analyse geographical sensor stability and continuous influence of dynamically variable point cloud mapping. This allows a thorough investigation of system boundaries and external influences on global depth sensor alignment. 

\end{document}